\documentclass{article}

\usepackage{arxiv}

\usepackage[utf8]{inputenc} % allow utf-8 input
\usepackage[T1]{fontenc}    % use 8-bit T1 fonts
\usepackage{hyperref}       % hyperlinks
\usepackage{url}            % simple URL typesetting
\usepackage{booktabs}       % professional-quality tables
\usepackage{amsfonts}       % blackboard math symbols
\usepackage{nicefrac}       % compact symbols for 1/2, etc.
\usepackage{microtype}      % microtypography
\usepackage{lipsum}
\usepackage{svg}
\usepackage{graphicx}
\graphicspath{ {./images/} }
\usepackage{amsmath} 
\usepackage{pdflscape}
\usepackage{xcolor}
\usepackage[normalem]{ulem}
\usepackage{comment}
\usepackage{longtable}
\usepackage{tabularx}
\usepackage{caption} % for more complex captioning
\usepackage{subcaption}
\usepackage{longtable}
\usepackage{booktabs}
\usepackage[utf8]{inputenc}
\usepackage{changepage}

\usepackage{array}
\usepackage{float}
\newlength{\extralength}
\newlength{\fulllength}
\newcolumntype{C}{>{\centering\arraybackslash}X}

\title{A Review of Transformer-Based Models for Computer Vision Tasks: Capturing Global Context and Spatial Relationships}

\author{Gracile Astlin Pereira\textsuperscript{*} and Muhammad Hussain \\[1ex]
\begin{minipage}[t]{0.90\textwidth}
\centering
\scriptsize Department of Computer Science, Huddersfield University, Queensgate, Huddersfield HD1 3DH, UK \\
\textsuperscript{*}Correspondence: U2292824@unimail.hud.ac.uk
\end{minipage}}

\begin{document}

\maketitle
\begin{abstract}
Transformer-based models have transformed the landscape of natural language processing (NLP) and are increasingly applied to computer vision tasks with remarkable success. These models, renowned for their ability to capture long-range dependencies and contextual information, offer a promising alternative to traditional convolutional neural networks (CNNs) in computer vision. In this review paper, we provide an extensive overview of various transformer architectures adapted for computer vision tasks. We delve into how these models capture global context and spatial relationships in images, empowering them to excel in tasks such as image classification, object detection, and segmentation. Analyzing the key components, training methodologies, and performance metrics of transformer-based models, we highlight their strengths, limitations, and recent advancements. Additionally, we discuss potential research directions and applications of transformer-based models in computer vision, offering insights into their implications for future advancements in the field.
\end{abstract}

% Keywords

\section{Introduction}
\label{sec:introduction}
Transformer architectures, initially pioneered for natural language processing (NLP) tasks, have experienced a remarkable surge in popularity within the realm of computer vision \cite{vaswani2023attentionneed,zahid2023lightweight}. The transformative success of transformer-based models in NLP applications, including language translation and sentiment analysis, has sparked considerable interest in leveraging these architectures for solving computer vision challenges \cite{wu2022pale}. Unlike conventional convolutional neural networks (CNNs), which process images hierarchically through layers of convolutions, transformers offer a novel approach by enabling direct interactions among different regions of the image \cite{vaswani2023attentionneed,hussain2023review}. This capability facilitates the capture of global context and spatial relationships, thereby revolutionizing the way computers understand visual data \cite{zongren2023focal,aydin2023domain}.

In this comprehensive review, we delve into the diverse landscape of transformer architectures tailored specifically for computer vision tasks. We examine how these architectures have evolved to address the unique requirements and challenges posed by visual data analysis. By harnessing the power of self-attention mechanisms and innovative design principles \cite{amini2021t6d}, transformer-based models have demonstrated remarkable capabilities in various computer vision domains, ranging from image classification to object detection \cite{liu2021swin} and semantic segmentation \cite{yin2022parotid}.

One of the key distinguishing features of transformer-based models is their ability to process the entire image or image patches simultaneously. Unlike CNNs, which rely on hierarchical feature extraction, transformers leverage self-attention mechanisms to capture global dependencies and contextual information in a single pass \cite{chen2024image}\cite{shi2023object}. This holistic approach not only improves the efficiency of information processing but also enables the models to effectively encode long-range dependencies and capture intricate spatial relationships within the image.

Throughout this review, we will explore several prominent transformer architectures tailored for computer vision tasks, including the Vision Transformer (ViT) \cite{dosovitskiy2020image}, DEtection TRansformer (DETR) \cite{carion2020end}, Spatially Modulated Co-Attention (SMCA) \cite{gao2021fast}, SWIN Transformer \cite{liu2021swin}, Anchor DETR \cite{wang2022anchor}, and DEformable TRansformer \cite{zhu2020deformable}. Each of these architectures offers unique advantages and innovations aimed at advancing the state-of-the-art in computer vision. We will delve into their underlying mechanisms, architectural designs, and performance characteristics, shedding light on their implications for the future of visual perception and understanding.

By providing a comprehensive overview of transformer-based approaches in computer vision, this review aims to offer insights into the current trends, challenges, and opportunities in the field. We will discuss the potential applications, limitations, and future directions of transformer architectures, as well as their impact on reshaping the landscape of computer vision research and development. Through a critical analysis of existing methodologies and emerging trends, we strive to provide valuable insights for researchers, practitioners, and enthusiasts interested in exploring the transformative potential of transformers in visual intelligence ~\cite{hussain2023child}.

\section{Transformer Architectures for Computer Vision}
\subsection{Vision Transformer (ViT)}
Vision Transformer \cite{dosovitskiy2020image} revolutionized the field of computer vision by introducing the transformer architecture directly to image classification tasks. Unlike traditional CNNs, which rely on hierarchical feature extraction through layers of convolutional operations, ViT takes a different approach. It breaks down input images into fixed-size patches, treating them as sequences of tokens, akin to words in natural language processing tasks. These image patches are then linearly embedded into feature vectors and fed into transformer layers for further processing \cite{yang2024stellar}.

\begin{figure}[H]
\centerline{\includegraphics[width=1\textwidth]{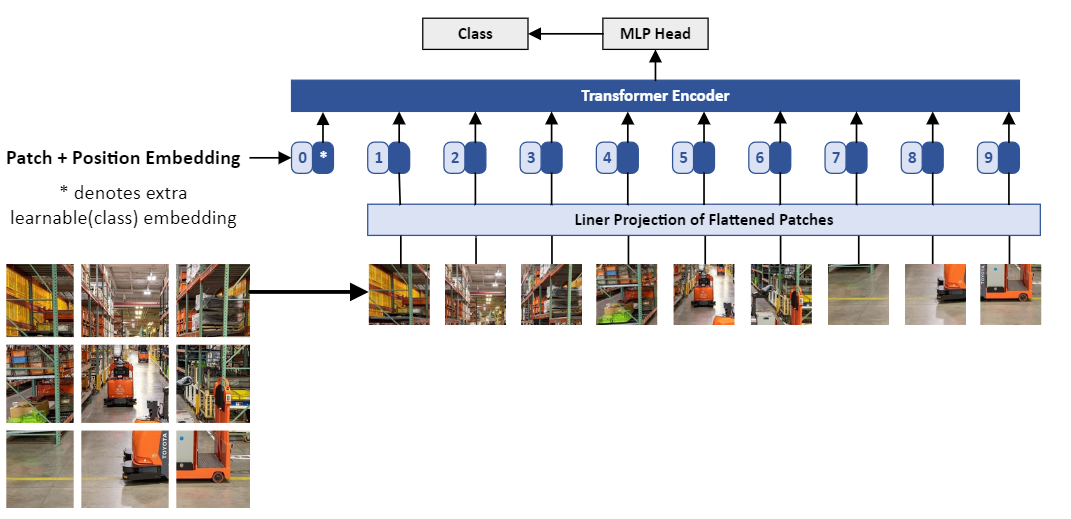}}
\caption{Vision Transformer Architecture}
\label{figure1}
\end{figure}

The core idea behind ViT is to leverage the self-attention mechanism of transformers to capture global dependencies and contextual information within images. By processing the entire image in one holistic view, ViT eliminates the need for handcrafted features or spatial hierarchies, offering a more flexible and adaptive approach to image understanding \cite{su2022mask}. Each patch in the input image interacts with all other patches through self-attention, allowing the model to attend to relevant image regions while ignoring distractions or noise \cite{he2023vhr}, as presented in Figure \ref{figure1}.

\subsubsection{Internal Architecture Breakdown:}
\textbf{Patch Embeddings:} The input image is divided into fixed-size patches, typically 16x16 or 32x32 pixels, forming a grid-like structure. Each patch is then linearly embedded into a lower-dimensional feature space using a learnable linear projection. This process converts each patch into a feature vector, which serves as the input to the transformer encoder \cite{yun2022patch}.

\textbf{Transformer Encoder Layers:} The patch embeddings are passed through multiple transformer encoder layers. Each encoder layer consists of self-attention mechanisms and feed-forward neural networks. In the self-attention mechanism, each patch attends to all other patches in the image, capturing global dependencies and contextual information. The feed-forward neural networks process the attended features, allowing the model to learn complex representations of the input image \cite{raganato2018analysis}.

\textbf{Positional Encodings:} Since transformers do not inherently encode spatial information, positional encodings are added to the patch embeddings to provide information about the spatial arrangement of patches within the image. These positional encodings can take various forms, such as sinusoidal functions or learned embeddings, and are added to the patch embeddings before feeding them into the transformer encoder layers \cite{zhang2023positional, chu2021conditional}.

\textbf{Multi-Head Attention Mechanism:} Within each transformer encoder layer, the self-attention mechanism is typically implemented as multi-head attention. This involves splitting the input feature vectors into multiple heads and computing separate attention scores for each head. The outputs of the attention heads are then concatenated and linearly transformed to produce the final output of the self-attention layer \cite{alamri2021multi, pedro2022assessing}.

\textbf{Feed-Forward Neural Networks:} After the self-attention mechanism, the attended feature vectors pass through a feed-forward neural network (FFNN) within each transformer encoder layer. The FFNN consists of two linear transformations separated by a non-linear activation function, such as GeLU (Gaussian Error Linear Unit). This allows the model to capture complex patterns and interactions between patches in the image \cite{bai2023multi, melas2021you}.

\textbf{Layer Normalization and Residual Connections:} Layer normalization and residual connections are applied after each sub-layer (self-attention and feed-forward neural network) within the transformer encoder layers. Layer normalization helps stabilize training by normalizing the activation of each layer, while residual connections facilitate gradient flow and alleviate the vanishing gradient problem \cite{yao2021leveraging, raghu2021vision, bao2023all}.

\textbf{Output Layer:} The output of the final transformer encoder layer is typically pooled across all patches to produce a global representation of the input image. This global representation is then passed through a classification head, which consists of one or more fully connected layers followed by a softmax activation function for image classification tasks \cite{bazi2021vision}. For other tasks like object detection, additional heads may be added to predict bounding boxes, class labels, and other relevant attributes.

In summary, the Vision Transformer (ViT) architecture combines the principles of transformers with innovative strategies for processing images, leading to state-of-the-art performance in computer vision tasks. By breaking down input images into patches and leveraging self-attention mechanisms, ViT captures global dependencies and contextual information effectively, paving the way for advancements in various vision tasks.

\subsection{DEtection TRansformer (DETR):}
DETR is a transformer-based model specifically designed for end-to-end object detection tasks, aiming to eliminate the reliance on traditional anchor boxes or region proposal networks \cite{carion2020end}. It introduces a novel approach where it directly predicts the bounding boxes and class labels for all objects in an image in a single pass. This paradigm shift offers a more streamlined and holistic solution to object detection tasks.

DETR employs a transformer encoder-decoder architecture tailored for object detection. The encoder processes the input image, extracting features and capturing global contextual information. On the other hand, the decoder generates object queries and predicts their attributes, such as bounding box coordinates and class labels \cite{sun2021rethinking}. This encoder-decoder architecture enables DETR to effectively encode the image information and decode it into meaningful object representations.

One of the key innovations introduced by DETR is its handling of the permutation problem inherent in object detection tasks. Unlike traditional approaches that rely on predefined correspondences between predictions and ground truth objects (e.g., anchor boxes), DETR tackles this problem using a novel bipartite matching loss mechanism \cite{ntinou2024multiscale}. This mechanism allows DETR to associate predicted objects with ground truth objects efficiently, enabling it to learn from object detection datasets without relying on explicit anchor boxes or region proposals.

\subsubsection{Internal Architecture Breakdown:}

\textbf{Transformer Encoder:} The input image is processed by a transformer encoder, which consists of multiple encoder layers. Each encoder layer incorporates self-attention mechanisms and feed-forward neural networks to capture global contextual information and extract features from the input image. The encoder transforms the input image into a set of feature vectors, which serve as the input to the decoder \cite{liu2021wb}.

\textbf{Transformer Decoder:} The transformer decoder generates object queries and predicts their attributes based on the encoded features provided by the encoder. It comprises multiple decoder layers, each of which includes self-attention mechanisms and feed-forward neural networks. The decoder iteratively refines the object predictions and refines their attributes, such as bounding box coordinates and class probabilities \cite{liu2021wb}.
 
\textbf{Bipartite Matching Loss:} DETR introduces a novel bipartite matching loss mechanism (Hungarian matching) to handle the permutation problem in object detection. This loss function associates predicted objects with ground truth objects in a bipartite matching manner, optimizing the alignment between predictions and ground truth annotations. By learning from the associations between predicted and ground truth objects, DETR effectively trains to predict accurate object bounding boxes and class labels \cite{jia2023detrs}.

\textbf{Training and Inference:} During training, DETR learns to predict object bounding boxes and class labels using ground truth annotations and the bipartite matching loss. During inference, the trained DETR model directly predicts object attributes for all objects in an image in a single pass, without the need for anchor boxes or region proposals. This end-to-end approach offers a more efficient and effective solution to object detection tasks \cite{chen2023enhanced}.

\begin{figure}[H]
\centerline{\includegraphics[width=1\textwidth]{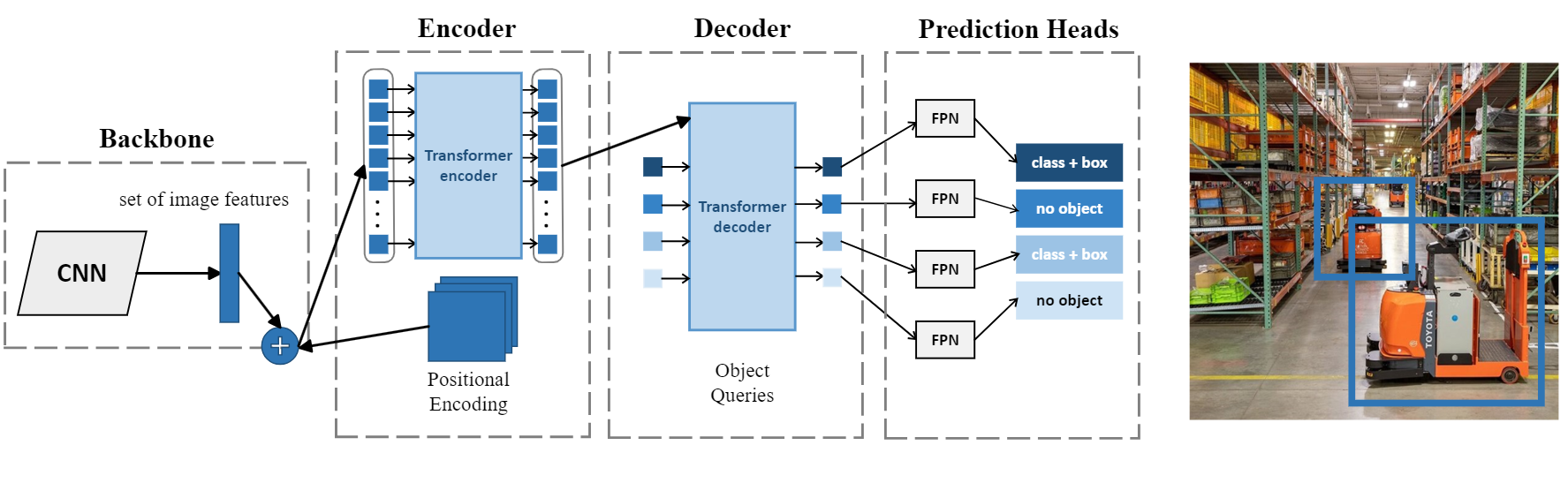}}
\caption{DEtection TRansformer Architecture}
\label{figure2}
\end{figure}

In summary, the DEtection TRansformer (DETR) architecture, presented in Figure \ref{figure2} redefines the approach to object detection by leveraging transformer-based models and introducing innovative mechanisms to address challenges inherent in traditional methods. With its encoder-decoder architecture and bipartite matching loss, DETR offers a promising solution for end-to-end object detection without relying on anchor boxes or region proposal networks.

\subsection{Spatially Modulated Co-Attention (SMCA):}
SMCA is an attention mechanism designed to enhance the interaction between image patches in a spatially modulated manner. By introducing spatially variant attention weights, SMCA enables the model to focus more on informative image regions while attending less to background noise \cite{gao2021fast}. This targeted attention mechanism aims to improve the performance of transformer-based models, particularly in tasks like object detection, by better capturing spatial relationships between image patches \cite{wu2024optimized}.

SMCA operates on the principle of modulating attention weights based on spatial information \cite{hu2020squeeze} within the image. Unlike traditional attention mechanisms that treat all image patches equally, SMCA dynamically adjusts attention weights to prioritize relevant image regions. By doing so, SMCA can effectively filter out background noise and focus on informative regions containing objects of interest.

One of the key objectives of SMCA is to capture spatial relationships between image patches more effectively. By modulating attention weights based on spatial information, SMCA enables the model to attend to neighboring patches that are spatially coherent, facilitating better understanding of object shapes and contexts within the image. This spatial awareness enhances the model's ability to perform tasks like object detection, where accurate localization and recognition of objects are crucial \cite{zhang2020deep}.

\subsubsection{Internal Architecture Breakdown:}

\textbf{Spatially Variant Attention Weights:} SMCA introduces spatially variant attention weights that are dynamically adjusted based on the spatial layout of the image. These attention weights determine the importance of each image patch during the attention mechanism, allowing the model to focus on relevant regions while suppressing background noise. The spatial variant nature of the attention weights enables SMCA to capture spatial relationships effectively \cite{malczewski2020emerging}.

\textbf{Modulation Mechanism:} The modulation mechanism in SMCA governs how attention weights are adjusted based on spatial information. This mechanism may involve incorporating spatial features or embeddings into the attention calculation process, enabling the model to learn spatially aware attention patterns. By modulating attention in this manner, SMCA ensures that the model attends to informative regions while suppressing distractions \cite{yang2023efficient}.

\textbf{Interaction Between Image Patches:} SMCA facilitates enhanced interaction between image patches by modulating attention weights. As the model processes the input image, attention is dynamically adjusted to prioritize spatially coherent patches that are likely to contain relevant information \cite{zhang2021csart}. This interaction between image patches enables the model to capture spatial relationships effectively, leading to improved performance in tasks like object detection.

\begin{figure}[H]
\centerline{\includegraphics[width=1\textwidth]{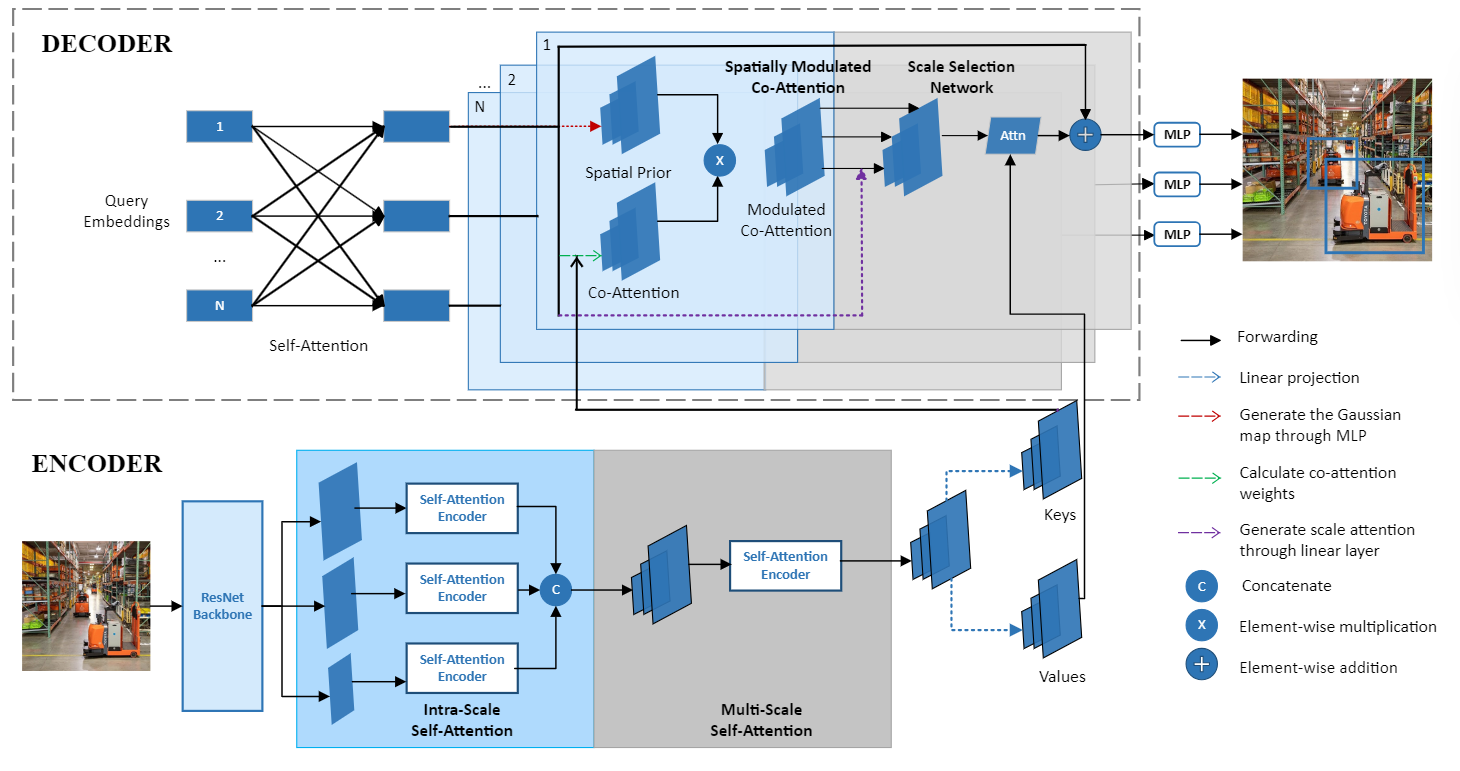}}
\caption{Spatially Modulated Co-Attention Architecture}
\label{figure3}
\end{figure}

\textbf{Task-Specific Adaptation:} SMCA can be adapted to different tasks and scenarios by adjusting its modulation mechanism \cite{9857671} and attention weighting scheme accordingly. For instance, in object detection tasks, SMCA may prioritize attention on regions containing objects while suppressing attention on background regions \cite{liu2023group}. This task-specific adaptation ensures that SMCA can effectively address the requirements of various computer vision tasks.

In summary, Spatially Modulated Co-Attention (SMCA) enhances the interaction between image patches by introducing spatially variant attention weights, as presented in Figure \ref{figure3}. By modulating attention based on spatial information, SMCA enables transformer-based models to better capture spatial relationships and focus on informative image regions, leading to improved performance in tasks like object detection.

\subsection{SWIN Transformer:}
SWIN Transformer introduces a hierarchical architecture that operates at multiple scales, aiming to capture both local and global contextual information efficiently \cite{liu2021swin}. It divides the input image into non-overlapping patches at different scales and processes them through transformer layers hierarchically. By allowing patches to interact across different levels of the hierarchy, SWIN Transformer enhances the model's ability to understand both fine-grained details and global context within the image.

SWIN Transformer addresses the challenge of capturing both local and global contextual information in large images effectively. Unlike traditional transformer-based models that process the entire image as a single sequence of patches, SWIN Transformer adopts a hierarchical approach. It divides the input image into non-overlapping patches at different scales, allowing the model to capture information at multiple levels of granularity \cite{shi2023face}.

One of the key innovations introduced by SWIN Transformer is its hierarchical processing strategy. Instead of processing all patches in parallel, SWIN Transformer hierarchically processes patches at different scales. This hierarchical processing enables the model to capture both local details and global context efficiently by allowing patches to interact across different levels of the hierarchy \cite{huang20223}.

\subsubsection{Internal Architecture Breakdown:}

\textbf{Hierarchical Patch Processing:} SWIN Transformer divides the input image into non-overlapping patches at different scales, forming a hierarchical structure. Each level of the hierarchy corresponds to patches of different sizes, enabling the model to capture information at multiple scales simultaneously. This hierarchical patch processing strategy facilitates the extraction of both local and global contextual information within the image \cite{cai2023hipa, li2022hst, sun2022hierarchical}.

\textbf{Transformer Layers:} At each level of the hierarchy, SWIN Transformer applies transformer blocks that apply self-attention within fixed-size, non-overlapping windows (for example, 7x7). This window-based self-attention confines computation to local regions of the feature map, improving efficiency \cite{liu2022swin}. Additionally, to enhance cross-window interactions, the model uses a shifted window mechanism where the window positions are shifted between successive blocks. This shifting enables information flow between adjacent windows. Each block also incorporates multi-head attention and a feed-forward network (MLP) to further refine the feature representations \cite{xu2021improved}.

\textbf{Cross-Level Interaction:} SWIN Transformer facilitates interaction by the use of shifted windows and patch merging layers. The shifted window mechanism allows for the interaction between previously isolated windows, enabling the model to capture long-range dependencies across different regions \cite{wang2023cross, wang2023stformer}. Furthermore, at each stage transition, patch merging layers downsample the feature maps and increase their depth, which aggregates information from previous levels and enhances cross-level interactions. This combination ensures that higher-level feature representations integrate contextual information from lower levels.

\textbf{Efficient Contextual Modeling:} By processing patches hierarchically and facilitating cross-level interaction, SWIN Transformer achieves efficient contextual modeling. The model can capture both fine-grained details and global context within the image while maintaining computational efficiency. This efficient contextual modeling enables SWIN Transformer to achieve state-of-the-art performance on various computer vision tasks, including image classification, object detection, and semantic segmentation \cite{themyr2023full, li2022contextual}. The shifted window approach further ensures that global context is integrated across multiple layers, balancing the need for efficiency with the ability to model diverse visual patterns.

\begin{figure}[H]
\centerline{\includegraphics[width=1.1\textwidth]{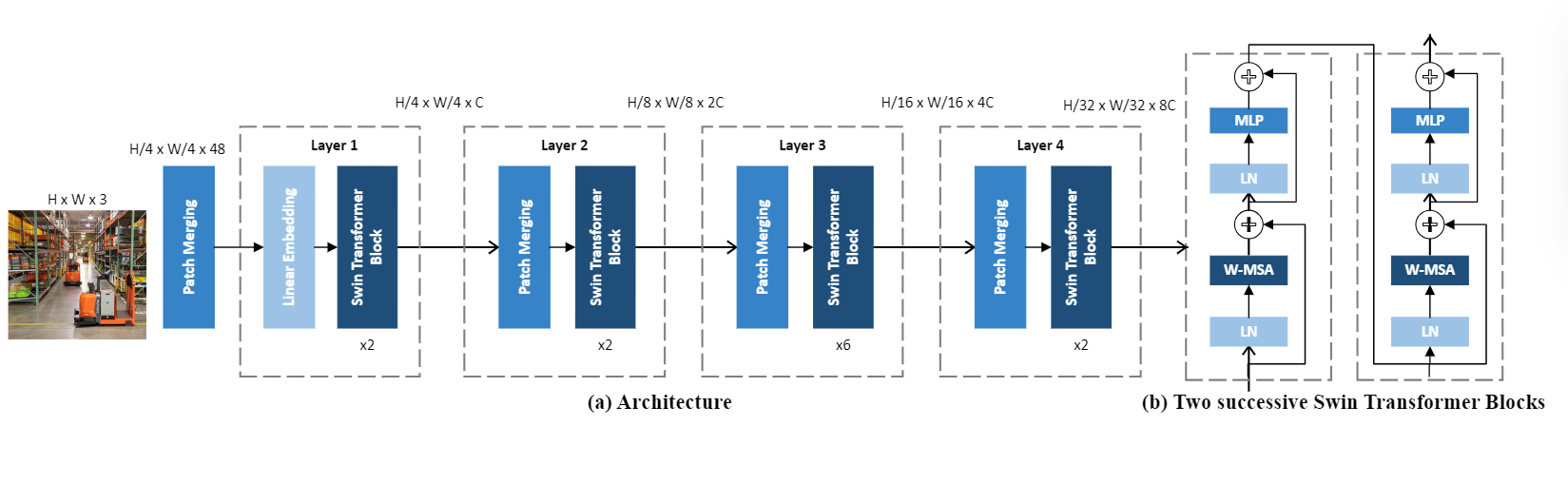}}
\caption{SWIN Transformer Architecture}
\label{figure4}
\end{figure}

In summary, SWIN Transformer introduces a hierarchical architecture that operates at multiple scales, as presented in Figure \ref{figure4}, allowing it to capture both local and global contextual information efficiently. By hierarchically processing patches and facilitating cross-level interaction, SWIN Transformer enhances its ability to understand complex visual scenes and make accurate predictions in various computer vision tasks.

\subsection{Anchor DETR:}

Anchor DETR combines the concepts of anchor-based object detection, akin to Faster R-CNN, with the transformer architecture. It utilizes predefined anchor boxes to guide the detection process, similar to traditional anchor-based methods \cite{wang2022anchor}. By leveraging the expressiveness of transformers for capturing contextual information and the efficiency of anchor-based methods for predicting object locations and attributes, Anchor DETR offers a promising approach to object detection tasks.

Anchor DETR bridges the gap between anchor-based object detection methods and transformer-based architectures, aiming to leverage the strengths of both approaches. In anchor-based methods like Faster R-CNN, predefined anchor boxes are used to generate region proposals, which are then refined to predict object locations and attributes. However, these methods may struggle to capture global contextual information effectively \cite{dai2022ao2}.

On the other hand, transformers have demonstrated remarkable success in capturing global dependencies and contextual information in various tasks. By processing the entire image as a sequence of patches and leveraging self-attention mechanisms, transformers can effectively model relationships between image regions. However, they may lack the efficiency and spatial precision offered by anchor-based methods for object localization.

\subsubsection{Internal Architecture Breakdown:}

\textbf{Transformer Backbone:} Anchor DETR employs a transformer backbone similar to other transformer-based models. The input image is divided into patches, which are then linearly embedded and processed by transformer encoder layers. This allows Anchor DETR to capture global contextual information and relationships between image patches effectively \cite{Sun2023FullyTD}.

\textbf{Anchor Boxes:} In addition to the transformer backbone, Anchor DETR incorporates predefined anchor boxes similar to traditional anchor-based methods. These anchor boxes serve as reference points for predicting object locations and attributes. By utilizing anchor boxes, Anchor DETR maintains the efficiency and spatial precision required for accurate object localization \cite{liu2022dab}.

\textbf{Object Detection Head:} Anchor DETR includes an object detection head responsible for predicting object locations and attributes based on the features extracted by the transformer backbone. This detection head leverages the contextual information captured by the transformer while utilizing anchor boxes to guide the detection process. By combining the strengths of transformers and anchor-based methods, Anchor DETR achieves accurate and efficient object detection \cite{Hong2022DynamicSR}.

\textbf{Training Strategy:} During training, Anchor DETR learns to associate predicted objects with ground truth annotations using a suitable loss function, such as a combination of classification and regression losses. This training strategy enables Anchor DETR to refine its predictions and improve its performance over time. By iteratively optimizing the detection head parameters, Anchor DETR learns to predict object locations and attributes accurately.

\begin{figure}[H]
\centerline{\includegraphics[width=1\textwidth]{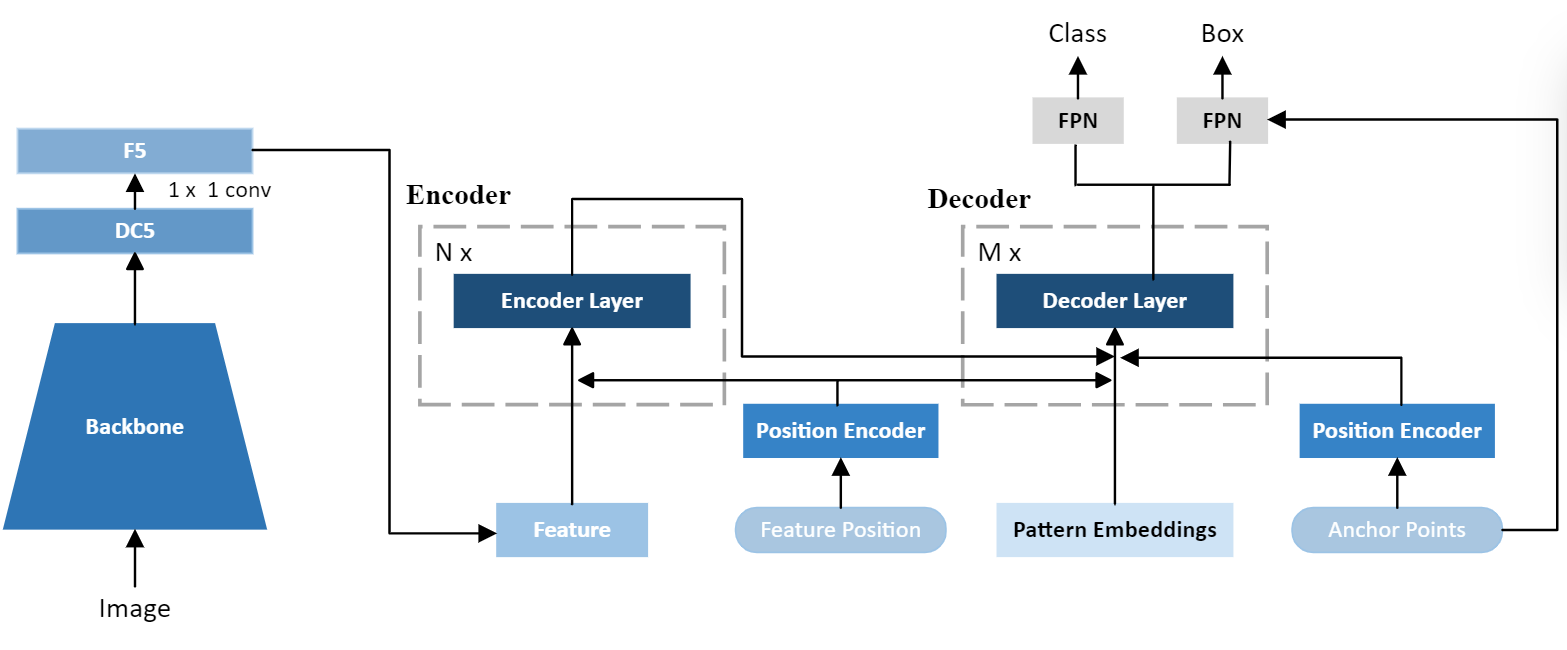}}
\caption{Anchor DEtection TRansformer Architecture}
\label{figure5}
\end{figure}

In summary, Anchor DETR, presented in Figure \ref{figure5} combines the concepts of anchor-based object detection with the transformer architecture, offering a promising approach to object detection tasks. By leveraging predefined anchor boxes to guide the detection process and capturing contextual information using transformers, Anchor DETR achieves accurate and efficient object localization and attribute prediction. This integration of anchor-based methods with transformers demonstrates the potential for advancing object detection capabilities in computer vision.

\subsection{DEformable TRansformer (DETR):}

DEformable TRansformer (DETR) is a variation of DETR that introduces deformable self-attention mechanisms to improve the modeling of spatial relationships between image patches. It allows the model to dynamically adjust the receptive field of each attention head based on the content of the image \cite{zhu2020deformable}. DEformable TRansformer aims to enhance the ability of transformer-based models to capture fine-grained spatial details and deformable object shapes in tasks like object detection.

DEformable TRansformer (DETR) builds upon the foundation of DETR by incorporating deformable self-attention mechanisms. Traditional self-attention mechanisms in transformers treat all spatial locations equally, which may limit their ability to capture fine-grained spatial details and deformable object shapes effectively. DEformable self-attention addresses this limitation by allowing the model to adaptively adjust the receptive field of each attention head based on the content of the image \cite{xia2022vision}.

DETR employs a transformer architecture, where the input image is divided into patches and processed by multiple transformer encoder layers \cite{isaac2022multi}. Each encoder layer consists of self-attention mechanisms and feed-forward neural networks, enabling the model to capture global contextual information and spatial relationships between image patches. However, traditional self-attention mechanisms may struggle to capture fine-grained spatial details and deformable object shapes accurately.

\subsubsection{Internal Architecture Breakdown:}

\begin{figure}[H]
\centerline{\includegraphics[width=1\textwidth]{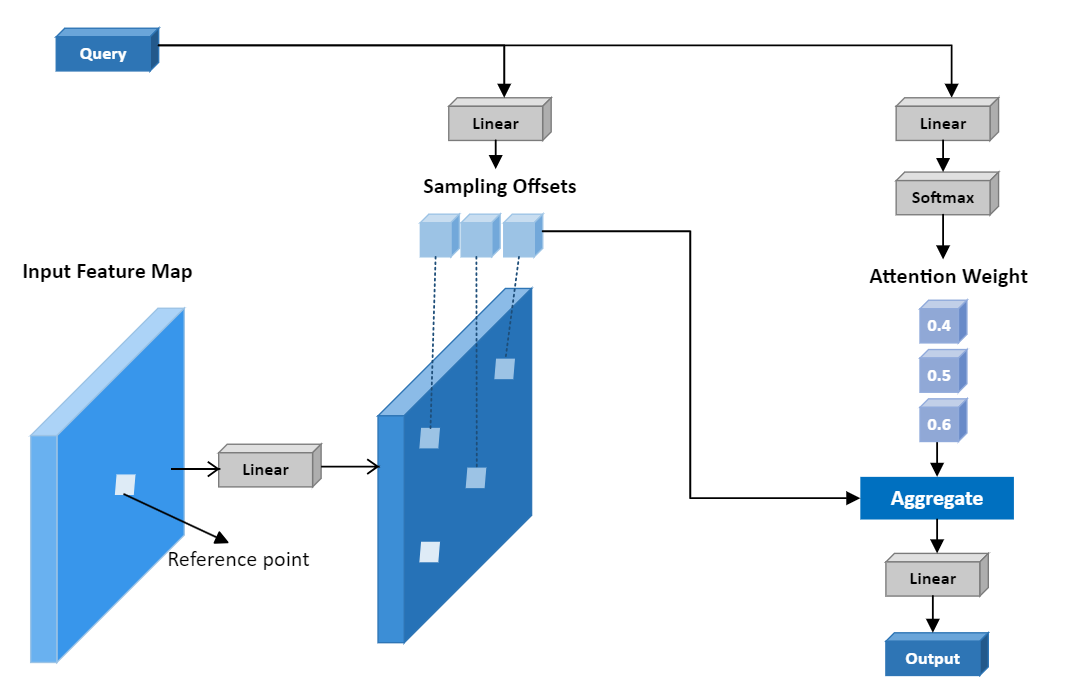}}
\caption{DEformable TRansformer Architecture}
\label{figure6}
\end{figure}

\textbf{Deformable Self-Attention Mechanisms:} DEformable TRansformer introduces deformable self-attention mechanisms that allow the model to dynamically adjust the receptive field of each attention head. Unlike traditional self-attention mechanisms, which treat all spatial locations equally, deformable self-attention mechanisms adaptively attend to informative regions of the image. This adaptive behavior enables the model to capture fine-grained spatial details and deformable object shapes effectively \cite{ma2021deformable}.

\textbf{Receptive Field Adjustment:} The deformable self-attention mechanisms in DEformable TRansformer dynamically adjust the receptive field of each attention head based on the content of the image. This adjustment allows the model to focus on relevant image regions while suppressing irrelevant or distracting information. By adaptively resizing the receptive field, DEformable TRansformer enhances its ability to capture spatial relationships and deformable object shapes accurately \cite{chen2021dpt}.

\textbf{Integration with Transformer Architecture:} DEformable TRansformer seamlessly integrates deformable self-attention mechanisms into the transformer architecture. The input image is still divided into patches and processed by multiple transformer encoder layers, as in traditional transformers. However, each encoder layer now incorporates deformable self-attention mechanisms, enabling the model to capture fine-grained spatial details and deformable object shapes more effectively \cite{liu2023deformer}.

\textbf{Training Strategy:} During training, DEformable TRansformer learns to adaptively adjust the receptive field of each attention head to optimize performance on the given task, such as object detection. This training strategy involves iteratively updating the parameters of the deformable self-attention mechanisms based on the model's predictions and ground truth annotations. By optimizing the receptive field adjustment process, DEformable TRansformer learns to capture spatial relationships and deformable object shapes accurately \cite{wei2023efficient}.

In summary, DEformable TRansformer (DETR), presented in Figure 6, enhances the modeling of spatial relationships between image patches by introducing deformable self-attention mechanisms. By dynamically adjusting the receptive field of each attention head, DEformable TRansformer improves its ability to capture fine-grained spatial details and deformable object shapes, leading to enhanced performance in tasks like object detection. This integration of deformable self-attention mechanisms with the transformer architecture demonstrates the potential for advancing spatial modeling capabilities in computer vision.

%%%%%%%%%%%%%%%%%%%%%%%%%%%%%%%%%%%%

\section{Capturing Global Context}
A core advantage of transformer-based models in computer vision lies in their capacity to comprehend global context effectively \cite{zhu2021vitt, guo2021sotr, ebert2023plg, hatamizadeh2023global}. Unlike convolutional neural networks (CNNs), which might encounter challenges in integrating information from distant image regions due to their hierarchical processing nature, transformers take a fundamentally different approach. They have the capability to process the entire image or image patches simultaneously, thereby enabling them to capture holistic information and long-range dependencies more efficiently.
\begin{table}[htbp]
    \centering
    \caption{Comparison of Transformers in Capturing Global Context}
    \label{tab:global-context}
    \begin{tabularx}{\textwidth}{|>{\raggedright\arraybackslash}p{2.5cm}|>{\raggedright\arraybackslash}p{2.5cm}|X|}
        \hline
        \textbf{Transformer} & \textbf{Global Context Capture} & \textbf{Description} \\
        \hline
        Vision Transformer (ViT) & Captures holistic information and long-range dependencies & ViT processes the entire image as a sequence of patches and utilizes self-attention mechanisms to assign varying levels of importance to different parts of the image. This enables ViT to understand complex visual scenes effectively, contributing to improved performance in tasks such as image classification and object detection \cite{yang2021focal}. \\
        \hline
        DEtection TRansformer (DETR) & Benefits from holistic view provided by transformers & DETR processes the entire image simultaneously and benefits from the holistic view provided by transformers. While its primary focus is on predicting object locations and attributes, DETR captures global dependencies and contextual information through self-attention mechanisms, enhancing its performance in object detection tasks \cite{lu2021end}. \\
        \hline
        Spatially Modulated Co-Attention (SMCA) & Prioritizes informative image regions & SMCA enhances the interaction between image patches in a spatially modulated manner, allowing the model to focus more on informative regions while attending less to background noise. While not directly focusing on global context, SMCA indirectly contributes to better scene understanding by prioritizing relevant image regions \cite{ni2019neuronal}. \\
        \hline
        SWIN Transformer & Captures both local and global contextual information efficiently & SWIN introduces a hierarchical architecture that operates at multiple scales, capturing both local and global contextual information effectively. By processing image patches hierarchically, SWIN integrates information across different levels of granularity, facilitating better understanding of complex visual scenes \cite{ghourabi2024advancing}. \\
        \hline
        Anchor DETR & Combines local context with global dependencies & Anchor DETR combines anchor-based object detection methods with transformers, allowing it to capture both local context and global dependencies effectively. By leveraging predefined anchor boxes and self-attention mechanisms, Anchor DETR improves performance in object detection tasks by considering both local and global information. \\
        \hline
        DEformable TRansformer & Improves modeling of spatial relationships and fine-grained details & DEformable TRansformer enhances the modeling of spatial relationships by introducing deformable self-attention mechanisms. These mechanisms allow the model to dynamically adjust the receptive field, facilitating better capture of fine-grained spatial details and deformable object shapes, leading to improved performance in tasks such as object detection \cite{shi2023daht}. \\
        \hline
    \end{tabularx}
\end{table}

The transformative power of transformers in capturing global context stems from their utilization of self-attention mechanisms. These mechanisms allow transformers to assign varying degrees of importance to different parts of the image based on their relevance to the task at hand \cite{vaswani2023attentionneed}. By dynamically adjusting attention weights during the processing of image features, transformers can effectively prioritize informative regions while suppressing irrelevant or noisy areas. This adaptive attention mechanism enables transformers to extract essential features from the entire image, ensuring that no relevant information is overlooked \cite{fan2024feature}.

The holistic view provided by transformer-based models facilitates a deeper understanding of the scene and context within an image. Unlike CNNs, which may focus primarily on local features and struggle to incorporate broader context, transformers can capture both local details and global relationships simultaneously. This comprehensive understanding of the image enables transformers to make more informed predictions and decisions, leading to improved performance across various computer vision tasks \cite{sahay2023importance}.

In tasks such as image classification and object detection, the ability to capture global context is particularly advantageous. It allows transformer-based models to consider the entire image when making predictions, rather than relying solely on local features or predefined regions of interest. This holistic perspective enables transformers to recognize complex patterns, infer semantic relationships between objects, and discern subtle visual cues that may be crucial for accurate classification or localization \cite{li_2022_ldetr}.

Overall, the capacity of transformer-based models to capture global context plays a pivotal role in enhancing their performance across a wide range of computer vision tasks. By processing the entire image in a holistic manner and leveraging self-attention mechanisms to prioritize relevant information, transformers can achieve superior performance compared to traditional CNN architectures. This ability to comprehend global context not only improves the accuracy and robustness of computer vision models but also opens up new opportunities for advancing the state-of-the-art in visual understanding and interpretation.

\section{Modeling Spatial Relationships}
Spatial relationships are fundamental aspects of visual understanding in computer vision tasks, particularly in tasks like object detection and semantic segmentation. Transformer-based models have introduced innovative techniques to effectively model spatial relationships between image patches, thereby enhancing their performance in these tasks \cite{carion2020end}.

One common approach employed by transformer-based models is the use of positional encodings. Positional encodings are additional input embeddings that encode spatial information into the input tokens of the model. By incorporating positional encodings, the model can distinguish between different locations in the image and understand the spatial layout of objects within the scene. These positional encodings are typically added to the input embeddings before feeding them into the transformer layers, allowing the model to learn spatial relationships along with other features \cite{chu2021conditional, zheng2021rethinking, chen2021simple}.

In addition to positional encodings, transformer-based models leverage cross-attention mechanisms to capture spatial dependencies between different regions of the image. Cross-attention mechanisms enable the model to attend to relevant regions of the image while processing each image patch, allowing for interactions between distant image regions. By dynamically adjusting attention weights based on the content of the image, the model can effectively capture spatial relationships and dependencies, even across long distances \cite{lu2024cross, huang2022transformer}.

Furthermore, transformer-based models may incorporate hierarchical architectures or multi-scale processing techniques to capture spatial relationships at different levels of granularity. By dividing the input image into non-overlapping patches at multiple scales, these models can capture both local details and global context, facilitating a more comprehensive understanding of spatial relationships within the image \cite{fang2021you, shou2022object}. This hierarchical processing enables the model to integrate information across different levels of abstraction, leading to improved localization accuracy and semantic understanding.

Overall, the effective modeling of spatial relationships is crucial for transformer-based models to perform well in computer vision tasks such as object detection and semantic segmentation.

\begin{table}[htbp]
    \centering
    \caption{Comparison of Transformers in Modeling Spatial Relationships}
    \label{tab:modeling-spatial}
   \begin{tabularx}{\textwidth}{|>{\raggedright\arraybackslash}p{2.5cm}|>{\raggedright\arraybackslash}p{2.5cm}|X|}
        \hline
        \textbf{Transformer} & \textbf{Modeling Spatial Relationships} & \textbf{Description} \\
        \hline
        Vision Transformer (ViT) & Employs positional encodings and self-attention mechanisms & ViT utilizes positional encodings to embed spatial information into the input tokens, enabling the model to distinguish between different locations in the image. Self-attention mechanisms further facilitate capturing spatial dependencies by allowing interactions between image patches. These techniques contribute to improved localization accuracy and semantic understanding in tasks such as object detection and semantic segmentation \cite{dosovitskiy2020image}. \\
        \hline
        DEtection TRansformer & Captures spatial relationships through self-attention mechanisms & DETR utilizes self-attention mechanisms to capture spatial relationships between image patches. By processing the entire image simultaneously, DETR can effectively model spatial dependencies and interactions, contributing to accurate object localization and semantic segmentation \cite{carion2020end}. \\
        \hline
        Spatially Modulated Co-Attention (SMCA) & Enhances interaction between image patches & SMCA introduces spatially variant attention weights, allowing the model to focus more on informative image regions while attending less to background noise. By enhancing the interaction between image patches in a spatially modulated manner, SMCA improves the model's ability to capture spatial relationships and dependencies effectively \cite{gao2021fast}. \\
        \hline
        SWIN Transformer & Hierarchical architecture captures spatial relationships efficiently & SWIN's hierarchical architecture enables it to capture spatial relationships efficiently by processing image patches at multiple scales. By dividing the input image into non-overlapping patches and processing them hierarchically, SWIN integrates information across different levels, facilitating better modeling of spatial dependencies and interactions \cite{liu2021swin}. \\
        \hline
        Anchor DETR & Leverages self-attention mechanisms for spatial modeling & Anchor DETR combines anchor-based object detection with transformers, leveraging self-attention mechanisms to model spatial relationships effectively. By considering interactions between predefined anchor boxes and image patches, Anchor DETR enhances its ability to capture spatial dependencies and improve localization accuracy \cite{wang2022anchor}. \\
        \hline
        DEformable TRansformer & Introduces deformable self-attention mechanisms for spatial modeling & DEformable TRansformer enhances spatial modeling by introducing deformable self-attention mechanisms. These mechanisms allow the model to dynamically adjust the receptive field based on the content of the image, facilitating better capture of spatial relationships and fine-grained details, leading to improved performance in tasks such as object detection and semantic segmentation \cite{zhu2020deformable}. \\
        \hline
    \end{tabularx}
\end{table}

\section{Training Methodologies and Performance Analysis}

Training transformer-based models for computer vision tasks often involves end-to-end training approaches, where the entire model is trained from scratch on large-scale datasets. This differs from traditional transfer learning approaches, where pre-trained CNNs are fine-tuned for specific tasks. Evaluation of transformer-based models in computer vision typically involves commonly used metrics such as accuracy, precision, recall, and F1-score. Additionally, performance on benchmark datasets such as ImageNet and COCO is used to assess the model's generalization capabilities and compare against existing methods.

Table 3 presents an overview of various transformer based implementations on vision tasks across a wide range of industrial applications from transmission line monitoring to surveillance operations and autonomous vehicles to agricultural automation.

\begin{longtable}{|p{1cm}|p{2.5cm}|p{2.5cm}|p{5cm}|p{3cm}|}
\caption{Overview of various studies and their details.} \\
\hline
\textbf{Refs.} & \textbf{Model} & \textbf{Application} & \textbf{Key Aspects} & \textbf{Results} \\
\hline 
\endfirsthead

\multicolumn{5}{c}{{\tablename} \thetable{} -- Continued from previous page} \\
\hline
\textbf{Refs.} & \textbf{Model} & \textbf{Application} & \textbf{Key Aspects} & \textbf{Results} \\
\hline 
\endhead

\hline \multicolumn{5}{r}{{Continued on next page}} \\ \hline
\endfoot

\hline \hline
\endlastfoot

\hline
\cite{garciamartin_2023_vision} & Vision Transformers (ViTs) & Vascular Biometric Recognition (VBR) & Pre-train ViTs on ImageNet-1k/21k dataset for feature extraction, fine-tune on four VBR variants using 14 datasets. Evaluate with TPIR and 75-25\% train-test splits. Introduce UC3M-CV3 dataset for wrist biometrics. & Accuracy: 96\% - 99.86\% \\ 
\hline
\cite{song_boosting} & Combination of vision transformers (ViT-B) and CNNs & Image Retrieval & R50+ViT-B/16, combines ResNet50 as the CNN stem and ViT-B/16 as the transformer encoder. Downsampling ratio of stem is 16. Embedding dimension throughout the encoder layers is D = 768, (ViT-B default). Proposed the design of Deep Token Pooling (DToP) & Accuracy: ROxf - 88.3\%; RPar - 91.9\% \\ 
\hline
\cite{gkelios_2021_investigating} & Vision Transformer descriptor & Content-Based Image Retrieval & The ViT descriptor, fully unsupervised and parameter-free, is easy to use without training or parameter fitting. The paper evaluates its performance using various similarity metrics and normalization techniques like L1-norm, L2-norm, and ROBUST normalization, demonstrating robustness and effectiveness compared to state-of-the-art local, global, and deep convolutional-based approaches across different datasets. & mAP: 64.7\% - 88\% \\ 
\hline
\cite{elnouby_2021_training} & Vision Transformers (ViTs) & Content-Based Image Retrieval & The methodology involves training Vision Transformers (ViTs) for image retrieval with a focus on contrastive learning. The study analyzes model components, compares feature extraction methods, especially the CLS token, and evaluates performance across objective functions like contrastive loss, Normalized Softmax, and Scalable Neighborhood Component Analysis (NCA). & Accuracy: 91.1\% \\ 
\hline
\cite{zhang_2023_vitaev2} & Vision Transformers -  ViTAE and ViTAEv2 models &  Object Detection, Instance Segmentation, Semantic Segmentation, and Pose Estimation & ViTAE and ViTAEv2 models incorporate intrinsic inductive biases (locality and scale invariance) via reduction and normal cells. ViTAEv2 features a multi-stage, divide-and-conquer design. Parallel PCM and attention modules in ViTAE extract local and global features, enhancing classification and computational efficiency. & Accuracy: 88.5\% \\
\hline
\cite{shamsabadi_2022_vision} & TransUNet (CNN + ViT backbone) & Crack Detection in Civil Infrastructure & The methodology involves training and evaluating TransUNet on datasets of crack images. The paper compares TransUNet with other models like U-Net and DeepLabv3+ in terms of accuracy and robustness to noise. It also discusses the advantages of ViTs in capturing global context and handling noise compared to CNN-based models. & Accuracy: 99.55\% \\ 
\hline
\cite{scheibenreif2023masked} & Vision Transformers & Land cover classification using hyperspectral remote sensing imagery & Utilise self-supervised masked image reconstruction for pre-training on large unlabeled hyperspectral datasets. Enhance transformer architecture with blockwise patch embeddings, spatial-spectral self-attention, and spectral positional embeddings. Study the impact on label efficiency, focusing on scenarios with limited labeled training data. & Accuracy: 82±0.2\% \\ \hline
\cite{gehrig_recurrent} & Recurrent Vision Transformers (RVTs) & Object detection with event cameras & The architecture includes convolutional layers, attention mechanisms, and LSTM cells across multiple stages. Spatial features are extracted via self-attention on grid partitions, while LSTMs handle temporal aggregation. Training uses mixed precision, a OneCycle learning rate schedule, and data augmentation. Performance is measured by mean average precision (mAP) on Gen1 and 1 Mpx datasets. & Gen1 - 47.2 mAP; 1 Mpx - 47.4 mAP \\ \hline
\cite{li_2022_mvitv2} & Multiscale Vision Transformer version 2 (MViTv2) & Object detection (COCO), Video recognition (Kinetics datasets) & Investigation of attention mechanisms (pooling, Swin, Hybrid window) for object detection.
Evaluation of positional embeddings and residual pooling connections.
Comparison between single-scale and multi-scale detection methods.
Training strategies and model configurations for video recognition tasks. & Accuracy: ImageNet - 88.8\%; Kinetics- 400 - 86.1\% \\ 
\hline
\cite{kim_2023_regionaware} & RO-ViT & Open-Vocabulary Object Detection, Image-Text Retrieval & Contrastive image-text pretraining strategies are explored, including different positional embeddings (e.g., standard, sinusoidal, cropped), contrastive losses (e.g., focal contrastive loss), and backbone architectures (e.g., ViT-B/16, ViT-L/16). Evaluation of frozen backbone features in detection fine-tuning.
Ablation studies on backbone learning rate ratio, downstream detector improvements, and scalability with model size and batch size.
Visualization of positional embeddings to compare learned representations. & AP$_r$ - 32.1 \\ \hline
\cite{fang_2023_unleashing} & MIMDET (Mask R-CNN with ViT) & Object detection (OD) and Instance Segmentation (IS) (COCO) & MIMDET adapts pre-trained ViT representations for object detection by introducing a ConvStem to replace the large-stride patchify stem, enabling pyramidal feature hierarchy. This hybrid ConvNet-ViT architecture uses ConvNet for early stages and ViT for later stages, employing random sampling and minimalist fine-tuning. & Small: OD - 51.7, IS - 46.1 mAP, Large: OD - 54.3, IS - 48.2 mAP \\ \hline
\cite{zhang_2023_a} & ViT-WSS3D & 3D object detection (SUN RGB-D, KITTI) & ViT-WSS3D employs a plain and non-hierarchical vision transformer to process point annotations and point clouds, generating pseudo-bounding boxes that can be used with any 3D detectors without additional modifications. & mAP@25: 52.4 - 58.8 \\ \hline
\cite{zhang_2023_efficient} & EIA-PVT & Oriented object detection in remote sensing images & Pyramid Positional Encoding (PPE) for scale-invariance inductive bias. Enhanced Inductive Encoders (EIEs) for reducing computational complexity and inducing global and local semantic relations. Angle Token Modulator (ATM) for embedding orientation knowledge into Transformer tokens. The architecture employs a hierarchical pyramid scheme and utilizes token sparsity to enhance efficiency while maintaining accuracy. & Accuracy: 65.8\% \\ \hline
\hline
\cite{ouyang_2024_deyo} & DEYO: DETR with YOLO & Object Detection & DEYO utilizes a progressive training strategy, consisting of two stages. In the first stage, the backbone and neck components are pre-trained on the COCO dataset. In the second stage, the model is fine-tuned using a query-based training approach. & Accuracy: 92.3\% \\
\hline
\cite{liu_2023_skip} & DETR (Detection Transformer) & Object Detection on Forest pests dataset & The methodology involves enhancing the DETR model with skip connections and a spatial pyramid pooling layer. Skip connections improve feature extraction, while the spatial pyramid pooling layer improves object query initialization. The model is trained using Python 3.9, Torch 1.11, and CUDA 11.4. & Accuracy: 65.7\% - 86.8\% \\
\hline
\cite{liu_2023_zeroshot} & DETR framework with semantics-aware attention mechanism  & Zero-Shot Object Detection (ZSD) in unconstrained images & The methodology involves integrating a semantics-aware attention mechanism into the DETR framework to address model bias towards seen classes. Additionally, a class-wise adaptive contrastive loss is developed to promote the learning of discriminative features while preserving semantic structural relationships between categories. & mAP50 = 19.5\% \\
\hline
\cite{li_2023_clsdetr} & CLS-DETR specifically modified from Sparse R-CNN & Object Detection & CLS-DETR introduces the Classification Information to Channel (C2C) mapping, channel multiplication, and learnable channel bias (LCB) modules. The Classification-Channel-Object Query mapping enhances object query and classification relations, accelerating convergence. Ablation experiments validate the effectiveness of each CLS-DETR component. & Accuracy: 45.3\% \\
\hline
\cite{hu_2023_emo2detr} & EMO2-DETR & Object Detection in Remote Sensing Imagery & The key components of EMO2-DETR include RBGM (Reassigned Bipartite Graph Matching) to handle high-quality negative samples, ISPH (Ignore Sample Prediction Head) for predicting high-quality negative samples, and reassigned Hungarian loss for parameter update. These components aim to improve the efficiency and accuracy of object detection, particularly in scenarios with varying object densities. & Accuracy: 76.23\% - 78.46\% \\
\hline
\cite{pu_rankdetr} & Rank-DETR & 3D Object Detection and Semantic Segmentation &  The core methodology revolves around incorporating rank-oriented designs to establish a more precise ranking order of predictions. Introduces a classification loss function that is aware of Generalized Intersection over Union (GIoU), effectively suppressing false positives and improving classification scores. Utilizes a high-order matching cost to push negative queries away from ground truth boxes, further enhancing detection performance. & mAP = 50.2\% \\
\hline
\cite{huang_dqdetr} & DQ-DETR (Dynamic Query Detection Transformer) with ResNet50 backbone & Aerial Image Object Detection & Utilizes a Categorical Counting Module (CCM) to estimate object count, employing classification for stability. Dynamically adjusts query count based on object density. Enhances query positional info for tiny object localization. Evaluated on AI-TOD-V2 with AP, APvt, and LRP metrics. Ablation studies favor classification over regression in CCM. & Accuracy: 94.6\%  \\
\hline
\cite{choi_2023_recurrent} & Recurrent DETR & Object Detection in Crowded Scenes & Recurrent DETR iteratively captures a larger prediction set. It uses Pondering Hungarian loss and Novelty Bias to halt computation dynamically. Analysing components removal impacts performance. Qualitative examples illustrate prediction process and detection patterns. & AP - 46.2\% \\
\hline
\cite{zeng_2024_arsdetr} & ARS-DETR with backbones ResNet-50 and Swin Transformer-T. & Oriented Object Detection in remote sensing and surveillance & Dynamically adjusts smoothing in angle classification, matching, and loss calculation processes based on the sensitivity of objects with different aspect ratios to angles.
Aligns features in DETR's decoder to improve detection accuracy.
Adopts a denoising strategy during training to enhance the model's robustness and performance.
Proposes AP75 as a more stringent performance metric for oriented object detection, arguing it better reflects the accuracy in high-precision scenarios. & ResNet-50 - AP50: 74.16\%, AP75: 49.41\%;  Swin Transformer-T - AP50: 75.47\%, AP75: 51.77\%  \\
\hline
\cite{li_2022_ldetr} & L-DETR with PP-LCNet as backbone  & Object Detection (Bounding boxes and Classification tasks) & L-DETR Architecture integrates PP-LCNet and a transformer-based model.
Transformer enhancements utilizes H-Sigmoid activation and group normalization.
Tests on various datasets to compare classification accuracy and bounding box detection with traditional DETR models.
Further, this paper highlights improvements in classification and bounding box detection, especially in resource-constrained scenarios. & bbox (IOU) = 0.0210 \\
\hline
\cite{misra_2021_an} & 3DETR & 3D object detection using point clouds & Use of non-parametric queries and Fourier positional embeddings is crucial for achieving good 3D detection performance, especially considering the irregular and sparse nature of point cloud data.
The paper introduces a set matching mechanism and a loss function that encourages a 1-to-1 mapping between predicted and ground truth boxes, thereby improving the robustness of the model.
3DETR is designed to be flexible and adaptable, allowing for variations in the number of queries and decoder layers at test time without retraining. & AP25 - 62.7\% and AP50 - 37.5\% \\
\hline
\cite{zheng_2021_endtoend} & Enhancement to DETR called ACT & Object Detection & ACT employs adaptive clustering to reduce redundancy in pre-trained DETR models.
It uses Exact Euclidean Locality Sensitive Hashing (E2LSH) for hashing.
Multi-Task Knowledge Distillation (MTKD) is applied to further improve ACT with a few-epoch fine-tuning.
The paper discusses hyper-parameter tuning for E2LSH, ablation studies, performance comparison with DETR and K-means clustering, and inference time and memory cost analysis. & AP - 43.1 \\
\hline
\cite{tang_2023_foreign} & YOLOX, incorporates Swin Transformer V2 and other modules like HSPP (Hybrid Spatial Pyramid Pooling) and RepVGGBlock for feature extraction and fusion. & Object Detection for transmission lines & Multi-head self-attention of the shifted window in Swin Transformer V2 is used for feature extraction.
The HSPP module is employed for expanding the receptive field and fusing multi-scale information.
RepVGGBlock is adopted to further improve feature extraction ability and detection accuracy. & mAP@50 - 96.7\%. \\
\hline
\cite{xuan_2023_swintransformerbased} & YOLOv5 with Swin-T & Object Detection in Remote Sensing Imagery & Use of CIOU for similarity index improvement in K-means clustering for anchor box generation.
Integration of modified Swin Transformer Block into YOLOv5 architecture.
Addition of Coordinate Attention to enhance detection accuracy of multi-scale targets.
Weighted bidirectional feature pyramid network for improved feature fusion.
Utilization of multi-scale feature extraction and fusion to enhance detection accuracy. & mAP: 74.7\% \\
\hline
\cite{jiang_2023_remote} & RAST-YOLO, based on YOLOv5 architecture & Object detection in remote sensing images & RA mechanism combined with Swin-T as the backbone for feature extraction to capture global background information and local details effectively.
C3D module for integrating deep semantic information with shallow semantic information to improve the detection accuracy of multi-scale targets and small targets.
ACmix Plus Detector for fully utilizing global and local information to output more accurate categories and target localization. & mAP50 - 69.8\%; mAP50:95 - 90.7\% \\
\hline
\cite{giroux_tfftradnet} &  Hierarchical Swin Transformer architecture, referred to as T-FFTRadNet & Object Detection tasks in radar data of autonomous vehicles & The paper emphasizes the use of Swin Transformers for object detection in radar data, highlighting their ability to retain linear complexity with respect to the input, which is crucial for deployment on low-power hardware.
Instead of traditional Fourier Transform methods, the paper proposes using fully connected layers that mimic FFTs, allowing the model to operate efficiently on raw ADC inputs without preprocessing. & Cars - 62.5\% AP; Trucks -  44.5\% AP  \\
\hline
\cite{shi_2024_small} &  STF-YOLO & Object detection of small tea buds for agricultural monitoring purposes & The paper explores various attention mechanisms but finds that their impact on small tea bud detection is not significant.
The model utilizes Depthwise Convolution and other parameter-efficient modules to reduce computation while preserving spatial information.
A novel loss function derived from Wise IOU is introduced to enhance the convergence and generalization capabilities of the YOLOv8 model. & mAP@50 - 89.4\%;  mAP50:95 - 71\% \\
\hline
\cite{ding_2023_dht} & Dynamic Vision Transformer (DHT), which incorporates hybrid window attention and DynamicToken Normalization (DTN) & Industrial Surface Defect Image Classification & Utilization of Dynamic Vision Transformer (DHT).
Integration of hybrid window attention and DynamicToken Normalization (DTN) to capture both global and local information effectively.
Evaluation on NEU surface defect dataset and DAGM dataset to verify the effectiveness of the proposed approach.
Comparative analysis with other normalization methods and models, including ConvNet-based models and Transformer-based models. & Accuracy: 95.9\% - 98.5\% \\
\hline
\cite{alrifaey_2022_hybrid} & Deep learning techniques that include SAE for automatic feature extraction and DEOA for optimal feature subset selection. LSTM is employed for time-series fault classification. & Fault Detection and classification in grid-connected PV systems & The methodology involves preprocessing the voltage signals using DWT, feature extraction using SAE, optimal feature selection using DEOA, and fault detection/classification using LSTM. The model's performance is evaluated based on accuracy rates, computational time, and robustness to noise. & Accuracy: Noiseless - 99.93\%, Noisy - >90\% \\
\hline
\cite{xu_2023_an} &  YOLOX and DeeplabV3+ & Monitoring photovoltaic panel integrity and detecting faults & Enhanced YOLOX model for precise identification and detection of anomalies on photovoltaic panels.
Replaced backbone network with Transformer-oriented PVT-v2 structure for better feature extraction.
Integrated CBAM attention mechanism to amplify pertinent features in extracted feature maps.
Optimized label allocation framework to rectify sample distribution discrepancies within the dataset.
Modified loss function to Dice loss function to improve segmentation precision.
Advanced segmentation technique building upon DeeplabV3+ algorithm, employing MobileNetV2 and adjusting downsampling factor. & mAP - 95.84\% \\
\hline
\cite{li_2024_photovoltaic} & Transformer-style model called M-E, which incorporates ELSA blocks & Fault Detection and diagnosis in photovoltaic panels & The methodology involves the integration of ELSA blocks into the MPViT architecture. These ELSA blocks enhance the attention to local features, improving the model's ability to detect faults in PV panels & Accuracy: IR: 90.7\% - 94.1\%, EL: 86.4\% - 88.5\% \\
\hline
\cite{xie_2021_dpit} & DPiT (With Image Transformers) & Defect detection in PV solar cells. & Employs two 3x3 convolutions for downsampling and additional layers for feature refinement.
Improves upon Shifted Windows-based MSA by modeling relations among all windows.
Integrates spatial and semantic information using convolutional layers and softmax function.
Adds positional information using depth-wise convolution with zero paddings.
Fuses low-level and high-level features using attention mechanism, with two MABs inserted at key stages. & Accuracy: 91.7\% \\
\hline
\cite{yin_2023_pvyolo} & PV-YOLO & PV panel fault detection & Integration of transformer-based feature extraction (PVTv2) with YOLO detection.
Utilisation of CBAM (Convolutional Block Attention Module) for feature fusion.
Label assignment strategy and loss function optimization for improved detection accuracy.
Experimentation involves pre-training the PVT model on a large dataset and fine-tuning for detection tasks.
Evaluation metrics include AP, mAP, model complexity measured in GFLOPs, precision, and recall. & mAP - 92.56\% \\
\hline
\cite{costa_2023_quantification} & ViT-$\mu$, Swin-T, and ResNet-50 for both binary and multi-class classification. & Quality control of solar cell manufacturing processes. & The methodology involves using machine learning models trained on labeled images of damaged and undamaged solar cells. The models are utilized to detect flaws such as cracks and cold solders and classify them. The extent of damage is quantified using segmentation algorithms. & ViT-$\mu$ - 97\% \\
\hline
\cite{wang_2023_defect} &  DefT (Hybrid Transformer for Defect Detection with CNN-based decoder via skip connections) & Surface defect detection in industrial scenarios & DefT combines transformer-based encoder modules with a CNN-based decoder via skip connections.
The model explores different types of position encoding methods to better encode local position information.
DefT incorporates a lightweight multi-pooling self-attention module to model multi-scale global contextual relationships explicitly.
The paper extensively evaluates the efficiency of DefT in terms of training efficiency, data efficiency, and computational complexity compared to other methods. & Accuracy: 97.15\% - 98.08\% \\
\hline
\cite{liu_2023_foreign} &  IDETR & Real-time monitoring and detection of foreign object shading such as leaves, bird droppings, and snow on PV modules & IDETR compensates for the deficiency of the Transformer in local modeling capability by introducing a convolutional attention module.
Transfer learning is employed to improve model stability and performance, with pre-training on a large-scale dataset and fine-tuning on a small dataset. & AP (s, m , l) - 0.066, 0.192, and 0.234 \\
\hline
\cite{shang_2023_defectaware} & DAT-Net that uses Transformer-based encoder & Surface Defect Detection (SDD) for industrial purposes & Transformer-based Encoder replaces traditional convolutional methods with Transformer architecture.
Defect-aware Module (DAM) enables the model to perceive and capture defect geometry and characteristics.
Graph Position Encoding (GPE) provides positional information through dynamic graph construction on tokens.
Experimental Setups include field experiments on blade defect and tool wear datasets to validate the method. & mIoU - 87.24 \\
\hline
\cite{hong_2023_classification} & Compact Convolutional Transformer (CCT) & Fault classification in PV systems & The paper trains a CCT model using Taguchi's method for heatmap generation in fault classification. PSO optimizes the model. They compare CCT to CNNs (AlexNet, ResNet50, etc.) and machine learning methods (SVM, Decision Tree, etc.) & Accuracy: 97.34\% \\
\hline
\cite{he_2023_dcmfafnet} & DCMF-AFNet & Detection of hot-spot faults in PV systems & The DCMF-AFNet model incorporates modules such as DCTM (Dynamic Contextual Task Module), BiMFF (Bi-branch Multi-scale Feature Fusion), and DT-Head (Dynamic Task Head) to improve detection accuracy. The network is designed to accurately detect hot-spot fault targets even in challenging inspection environments characterized by motion blur and noise disturbance. & Accuracy: 87.3\%  \\
\hline
\cite{kang_2023_vision} & Vision Transformer (VIT) & The direct prediction of PV energy output & The methodology involves using the VIT model and incorporating auxiliary PV sensor information into the network input to improve prediction accuracy. The study also conducts ablation experiments to determine the optimal number of layers for feature extraction in the encoder & 300s - MSE: 9245.2, MAPE: 8.94\%; 600s MSE: 23763, MAPE: 16.45\%. \\
\hline
\cite{chen_2023_hatransformer} & Mask R-CNN with HA-Transformer Backbone & Object Detection & The methodology of the study involves the development and evaluation of the HA-Transformer architecture for object detection tasks. Bridge tokens improving model understanding of spatial relationships. Additionally, a transition module ensures efficient token exchange between neighboring windows, while the SS module reduces memory usage and computational costs by leveraging channel shifting techniques. Training and evaluation with the Mask R-CNN framework, optimizing hyper parameters like learning rate and weight decay & AP - 64.7\% \\
\hline
\cite{li_2023_mask} & Mask DINO incorporates a Swin Transformer backbone & Instance segmentation (IS), panoptic segmentation (PS), and semantic segmentation (SS) & Mask DINO unifies object detection and image segmentation tasks into a single model, leveraging the Transformer architecture. The model utilizes a Swin Transformer backbone for feature extraction. Mask DINO is pre-trained on the Objects365 dataset for better generalization. The paper conducts extensive ablation studies to analyze the effectiveness of different components, such as query selection, mask-enhanced anchor box initialization, feature scales, decoder layer number, and matching methods. & IS - 54.5 AP, PS - 59.4 PQ, SS - 60.8 mIoU \\
\hline
\cite{zhigang_2021_updetr} & DETR & Object Detection (OD) (in PASCAL VOC and COCO), One-shot Detection (OSD) (CNP in training data), Panoptic Segmentation (by adding a mask head above decoder outputs) & Introduces the pretext task of random query patch detection to pre-train transformers, focusing on spatial localization and feature discrimination.
Utilizes unsupervised pre-training with the pretext task before fine-tuning on specific detection tasks. Implements feature reconstruction and frozen CNN backbone to preserve feature discrimination for classification tasks. & OD: 42.8 - 56.1 AP, PS: 44.5 PQ, OSD - 61.2 AP \\
\hline
\cite{baek_2023_swin} & Swin Transformer on YOLOv3 backbone, enhanced by Few-Shot Learning. & Fire detection to predict and monitor fire occurrence in real-time. & Utilised 35,159 fire prediction images from AI hub, at 50\%-10\%-40\%, train/validation/test split.
Implemented Swin Transformer with a shifted window mechanism for hierarchical transformations and reduced information loss.
Integrated with YOLOv3 and Feature Pyramid Network (FPN) to handle multi-scale features and improve real-time detection.
Employed meta-learning for efficient learning with small data, using a distance-based feature extractor to improve small object detection. & mAP - 51.2, accuracy - 70.84\% \\
\hline

\end{longtable}

\section{Discussion}

The study and analysis presented in this paper highlight the transformative impact of transformer-based models on computer vision tasks, particularly in their ability to capture and process global context more effectively than traditional convolutional neural networks (CNNs). Transformers use self-attention mechanisms to process entire images or image patches all at once, enabling them to capture comprehensive information and understand long-range relationships. This is different from convolutional neural networks (CNNs), which typically process information in a hierarchical manner and may have difficulty integrating data from distant regions of an image.

The comparison of various transformer-based models, demonstrates the diversity of approaches employed by different models to capture global context. For instance, the Vision Transformer (ViT) excels in understanding complex visual scenes by treating images as sequences of patches and applying self-attention to assign importance to different parts of the image. Similarly, models like DEtection TRansformer (DETR) and SWIN Transformer effectively leverage global context to enhance object detection and scene understanding. Additionally, models such as the Spatially Modulated Co-Attention (SMCA) and DEformable TRansformer show how transformers can adaptively prioritize informative regions and capture fine-grained spatial details, further contributing to their superior performance in tasks such as object detection and semantic segmentation.

A key takeaway from the analysis is the adaptability of transformer-based models in capturing both local and global contextual information. This adaptability is not only achieved through the use of self-attention mechanisms but also through innovations like positional encodings, cross-attention mechanisms, and hierarchical architectures. These techniques allow transformers to effectively model spatial relationships, a critical aspect of visual understanding that is essential for tasks requiring precise localization and interpretation of objects within an image.

Moreover, the ability of transformers to dynamically adjust attention weights based on the content of the image underscores their capacity to prioritize relevant information and suppress noise. This leads to a more robust and accurate interpretation of visual data, which is especially important in complex and cluttered scenes where discerning subtle visual cues is crucial for task success.

\section{Conclusion}

In conclusion, transformer-based models have brought a significant shift in the landscape of computer vision, offering advantages that surpass those of traditional CNN architectures, particularly in capturing global context and modeling spatial relationships. The use of self-attention mechanisms, positional encodings, and hierarchical architectures allows these models to process entire images holistically, integrating information from both local and global perspectives. This capability not only enhances the accuracy and robustness of transformer-based models but also positions them as a promising direction for future advancements in visual understanding and interpretation.

As the field of computer vision continues to evolve, the insights gained from this study suggest that transformer-based models will play a pivotal role in pushing the boundaries of what is achievable. Their ability to handle complex visual scenes, capture long-range dependencies, and model intricate spatial relationships makes them well-suited for a wide range of applications, from image classification and object detection to more advanced tasks like semantic segmentation and scene understanding. Future research may further explore the optimization of these models ~\cite{hussain2023and}, addressing challenges such as computational efficiency and scalability, to fully realize their potential in real-world applications.

%%%%%%%%%%%%%%%%%%%%%%%%%%%%%%%%%%%%%%%%%%
\begin{adjustwidth}{-\extralength}{0cm}

\bibliographystyle{unsrt}  % Changes bibliography style to unsorted
\bibliography{references}  % This points to the filename of your BibTeX file without the .bib extension

\end{adjustwidth}
\end{document}